\documentclass[robotics,article,submit,moreauthors,pdfLaTeX]{Definitions/mdpi} 
\firstpage{1} 
\pubvolume{1}
\issuenum{1}
\articlenumber{0}
\pubyear{2024}
\copyrightyear{2024}
\datereceived{ } 
\daterevised{ } 
\dateaccepted{ } 
\datepublished{ } 
\hreflink{https://doi.org/} 
\usepackage[utf8]{inputenc}
\usepackage{amsmath,amssymb,amsfonts}
\usepackage{bm}
\usepackage{url}
\usepackage{graphicx,graphics}
\usepackage{multirow}
\usepackage{booktabs}
\usepackage{textcomp}
\usepackage{tabularx}
\usepackage{relsize}
\usepackage[flushleft]{threeparttable}
\usepackage{rotating}
\usepackage{comment}
\usepackage{soul}
\usepackage{tikz}
\usepackage{algpseudocode}  
\usepackage{subcaption}
\usepackage{algorithm2e}

\Title{Vision-Guided Grasp Planning for Prosthetic Hands in Unstructured Environments}

\TitleCitation{Title}

\Author{Shifa Sulaiman $^{1}$\orcidA{}*, Akash Bachhar $^{2}$, Ming Shen $^{3}$, and Simon Bøgh $^{4}$ }

\address{%
$^{1}$ \quad Control and Automation Section, Department of Electronics Systems, Aalborg University, Denmark; \\
$^{2}$ \quad Department of Mechanical Engineering, National Institute of Technology,Durgapur, India; \\
$^{3}$ \quad The Technical Faculty of IT and Design
Antennas, Propagation and Millimeter-Wave Systems, Department of Electronics Systems, Aalborg University, Denmark; \\
$^{4}$ \quad Control and Automation Section, Department of Electronics Systems, Aalborg University, Denmark;
}

\corres{Correspondence: ssajmech@gmail.com; }

\abstract{Recent advancements in prosthetic technology have increasingly focused on enhancing dexterity and autonomy through intelligent control systems. Vision-based approaches offer promising results for enabling prosthetic hands to interact more naturally with diverse objects in dynamic environments. Building on this foundation, the paper presents a vision-guided grasping algorithm for a prosthetic hand, integrating perception, planning, and control for dexterous manipulation. 
A camera mounted on the set up captures the scene, and a Bounding Volume Hierarchy (BVH)-based vision  algorithm is employed to segment an object for grasping and define its bounding box. Grasp contact points are then computed by generating candidate trajectories using Rapidly-exploring Random Tree Star (RRT*) algorithm, and selecting fingertip end poses based on the minimum Euclidean distance between these trajectories and the object’s point cloud. Each finger’s grasp pose is determined independently, enabling adaptive, object-specific configurations. Damped Least Square (DLS) based Inverse kinematics solver is used to compute the corresponding joint angles, which are subsequently transmitted to the finger actuators for execution. This modular pipeline enables per-finger grasp planning and supports real-time adaptability in unstructured environments. The proposed method is validated in simulation, and experimental integration on a Linker Hand O7 platform.}


\begin{document}



\section{Introduction}

A vision algorithm is significant in robotic applications as it enables robots to perceive, localize, and interpret their environment, providing the essential spatial awareness needed for accurate manipulation, adaptive motion planning, and safe human–robot interaction \cite{1_new}, \cite{2_new}. Vision-guided grasping has become a cornerstone of robotic manipulation, enabling systems to perceive and interact with unstructured and dynamic environments \cite{1}. For prosthetic hands, vision offers a pathway to restore more natural and autonomous object handling by providing rich spatial information about object shape, pose, and occlusions \cite{2}. Despite these advantages, practical vision-based grasping for prosthetics faces several intertwined challenges: noisy and partial point clouds from compact hand-mounted sensors, the need for fast real-time processing on embedded hardware, and the difficulty of converting geometric perception into kinematically feasible finger motions for highly articulated hands.

Many contemporary approaches simplify perception by using coarse object representations such as 2D bounding boxes, centroids, or single-view meshes \cite{3}. These simplifications reduce computational load but sacrifice geometric fidelity, which degrades grasp accuracy on irregularly shaped, deformable, or partially occluded objects. Equally limiting is the common practice of planning grasps for the hand as a single rigid entity. Treating the hand monolithically ignores per-finger kinematic constraints and the potential benefits of independent finger placement, reducing adaptability when objects require nonstandard contact distributions or when individual fingers must negotiate local obstacles.

To address these shortcomings, we propose a modular, per-finger grasping pipeline tailored to a prosthetic platform (LinkerHand O7) that tightly coupled perception, planning, and low-level control. Our perception module used a Bounding Volume Hierarchy (BVH) representation along with Axis-Aligned Bounding Boxes (AABB) built from hand-mounted point cloud data to produce compact, tight-fitting 3D object models that preserved local geometry and handle partial observations robustly. Using BVH enabled fast collision queries and distance computations that scaled to complex surfaces while remaining tractable for embedded processors. Instead of precomputing a small set of canonical grasp templates, we generated candidate fingertip trajectories online using RRT* algorithm, sampling trajectories that accounted for each finger’s workspace and joint limits. Inverse kinematic solutions are obtained using Damped Least Square (DLS) approach.
Major contributions of the paper are as follows:

\begin{itemize} \item \textbf{Per-finger, trajectory-aware grasp planning.} We computed each finger's end pose independently by generating candidate fingertip trajectories and selecting contacts that minimize trajectory-to-surface (Euclidean) distance, enabling adaptive, kinematically feasible, and object-specific grasps. \item \textbf{BVH-AABB perception for tight 3D segmentation.} A BVH using AABB produced compact, tight-fitting 3D object models from hand-mounted point clouds, improving contact localization and supporting fast collision and distance queries suitable for real-time planning. \item \textbf{End-to-end vision-to-actuation pipeline.} Vision-derived fingertip poses are converted to joint commands through inverse kinematics and transmitted to a hand actuators, providing a direct, low-latency path from perception to execution. \item \textbf{Modular, scalable architecture with validation.} The pipeline's modular design supported easy replacement or extension of perception, planning, or control modules and is validated in simulation and experimentation on a LinkerHand O7 prototype, showing improved grasp precision. \end{itemize}

\section{Background}
Robotic and prosthetic hands increasingly integrate perception and control to achieve reliable manipulation in everyday environments \cite{4}, \cite{5}. Mobile manipulators leverage vision systems to perceive their surroundings, enabling autonomous navigation and object interaction in dynamic, unstructured environments \cite{6}, \cite{7}. Efficient geometric representations and fast collision/distance queries enable planners to evaluate candidate motions against sensed geometry in real time. Per-finger planning that accounts individual kinematics improves adaptability and contact precision for irregular or partially occluded objects. Modular architectures facilitate incremental development and simplify integration with diverse prosthetic hardware.


Morales \textit{et al.} \cite{8} proposed a vision-guided pipeline that extracted object contours and computed two and three fingers planar grasps using grasp-region segmentation and curvature cues, explicitly constraining and scoring candidate grasps. 
The algorithmic stack included contour extraction, curvature-based segmentation into grasp regions, combinatorial force-closure tests for finger sets, geometric projection for grasp focusing, and kinematic feasibility checks coupled with heuristic quality aggregation. The work addressed practical online grasp synthesis for unknown objects under real-robot kinematic constraints. However it is limited to planar/2D assumptions without fully exploiting 3D object geometry or concavities, and lacks automatic tuning and broad statistical validation.
Saxena \textit{et al.} \cite{9} integrated learned visual grasp-point prediction with stereo-based scene modeling and a collision-tolerant planning pipeline to grasp novel objects in clutter. The method coupled a learned grasp classifier, fixed-background priors for scene reconstruction, and a probablistic road map (PRM) variant that tolerated spurious obstacle points, combined with pragmatic pre-shape and close-to-touch execution heuristics. The work showcased improved robustness in cluttered scenes and noisy sensing during practical tasks. However, the learned models are trained on a narrow set of object classes. Execution depended on handcrafted pre-shapes and lacked full closed-loop dexterous control.

Fang \textit{et al.} \cite{10} designed and evaluated a compact dual-modal vision-based fingertip sensor that sensed both surface texture (via a reflective membrane deformation) and distributed 3D contact forces (via tracked marker displacements), using a small camera, printed marker arrays, an MLP force estimator, and texture classification with rotationally invariant LBP features. The sensor enabled high-resolution tactile cues in a replaceable fingertip form factor and is validated in calibration and handheld tests, yet its integration into closed-loop dexterous control remained at the demo stage and questions about long-term durability and complex contact geometries are still to be addressed.
Cheng \textit{et al.} \cite{11} presented a dense, multi-scale deep detector for rotated grasp-box prediction with parallel grippers. It combined a ResNet50 backbone, dense Feature Pyramid Network (FPN), and a two-stage anchor-based detector. The model is validated on Cornell and Jacquard datasets and tested on a Franka robot with 51 unseen objects in clutter. It improved grasp-pose accuracy and multi-scale robustness for two-finger grippers. The method is not built to support multi-finger hands, tactile sensing, or closed-loop grasp refinement.
Fuentes \textit{et al.} \cite{12} proposed a vision–tactile pipeline for planar precision grasps using a four-finger hand. They extracted contours and normals to define a grasp-quality objective, optimized with a continuous genetic algorithm. Execution used uncalibrated visual servoing and tactile force feedback. The method enabled calibration-free planning and robust precision grasping in 2D. It is limited by planar assumptions, high optimization cost, and poor scalability to full 3D dexterous tasks.


Karaali \textit{et al.} \cite{13} evaluated eight Vision-Language Models on a 34-image prosthesis-view dataset. They tested whether a single VLM could predict structured JSON grasp parameters, including object identity, shape, orientation, dimensions, and grasp type. The models showed strong performance on categorical outputs like names and shapes, but numerical predictions varied in accuracy. Latency and computational cost are also differed significantly across models. The study suggested that VLMs are promising for high-level grasp planning. However, safe prosthetic use still required larger datasets, closed-loop validation, fine-tuning, and better handling of latency and output variance.
Wilson \cite{14} introduced TAFFI, a lightweight pinch detection method for desktop interaction. It segmented the hand from the background, detected a hole formed by a pinch gesture, and fits an oriented ellipse to estimate position, orientation, and scale. The system supported one and two hand gestures for tasks like cursor control and zooming. TAFFI is simple and effective in constrained setups but relied on clean segmentation, specific hand poses, and a narrow sensing range. It worked best as a quick intent signal, not as a full solution for precise manipulation.
Smith and Papanikolopoulos \cite{15} developed an eye-in-hand visual servoing system for grasping static or slow-moving rigid objects. Their pipeline used coarse-to-fine feature selection, adaptive inverse-depth (1/Z) estimation, and a MIMO control law to guide the robot without requiring calibration. Experiments on the MRVT platform confirmed accurate alignment and successful grasping. However, the method relied on hand-engineered features, incurred high computational cost, and was sensitive to feature loss and singularities. It also struggled to scale to faster object motions or more varied object types.


Park and Kim \cite{16} developed the HRI hand, an open-source, low-cost anthropomorphic five-finger robot hand designed for collaborative manipulation research. The platform included full CAD models, firmware, and ROS packages under an MIT license, supporting modular integration and reproducibility. Each finger used two underactuated four-bar linkages driven by linear actuators, with an additional motor for thumb abduction/adduction. The system ran on an STM32 microcontroller with Bluetooth and integrates with ROS through URDF and rviz. They validated the design through fingertip force and speed measurements, along with grasp demonstrations. The hand showed promising performance for basic manipulation tasks and offers a flexible foundation for further research. However, the current implementation lacked tactile sensing, force or impedance control, and closed-loop dexterous manipulation. These gaps limited its use in more advanced tasks requiring fine control and robust perception.
Shi \textit{et al.} \cite{17} developed a vision-based system to classify objects into four grasp types—cylindrical, spherical, tripod, and lateral using RGB-D input. Their goal is to simplify prosthetic hand control and reduce reliance on EMG signals. They introduced a new RGB-D dataset of daily objects captured under varied poses and lighting conditions, and demonstrated a semi-autonomous Vision-EMG control strategy. The system used CNN architectures for both mono-modal (grayscale) and bimodal (grayscale + depth) inputs. They conducted offline generalization tests and compared online prosthesis control performance against a coding-EMG baseline. Results showed improved grasp-type inference accuracy, especially when depth data is included in the dataset. The approach is not able to handle cluttered scenes or multi-object selection. It also lacked integration with eye-tracking or user intent signals, and its robustness to occlusions and different camera setups in real-world use remains untested.



Ficuciello \textit{et al.} \cite{18} demonstrated a synergy-based learning framework for grasping with an anthropomorphic hand-arm system. Their method combined neural network-based imitation learning, which mapped object features to low-dimensional synergy parameters, with model-based policy search using path integral techniques (PI2) in a reduced synergy space. Vision is used to estimate object shape and pose, with semantic recognition via RANSAC fitting to sphere and cylinder primitives from RGB-D point clouds. The approach used postural synergies (PCA eigengrasps) to reduce dimensionality and improve sample efficiency in high-DoF grasp learning. Neural networks provided initial grasp estimates, and policy search refined them for stability. This integration of vision and synergies enabled learning stable grasps on unknown objects with fewer samples. However, the method is limited in handling simple object geometries. 
Sayour \textit{et al.} \cite{19} presented a practical, real-time grasping pipeline for unknown objects using an eye-in-hand RGB-D camera. The system combined depth inpainting, visual servoing, and a lightweight GG-CNN grasp predictor with a MATLAB/Simulink controller running directly on a Barrett WAM arm, without relying on an internal PC. It emphasized modularity, industrial applicability, and multi-view grasp generation. Their methods included RGB-D preprocessing with bilateral filtering and inpainting, image-based visual servoing to center the object, and a fully convolutional GG-CNN to predict grasp quality, angle, and width. Post-processing refined grasp width using object gradients. 
However, The method failed in optimizing grasp approach kinematics across multiple views, showed limited generalization to novel object classes, and lacked tactile or force feedback integration for different grippers.
Markovic \textit{et al.} \cite{20} introduced a semi-autonomous prosthesis control system that combined stereovision grasp decoding with simple myoelectric triggers and augmented reality (AR) feedback. The system automatically selected and preshaped a dexterous prosthetic hand based on object geometry, while the user can fine-tune the grasp using low-bandwidth EMG signals. AR glasses provided visual feedback, acting as artificial proprioception to guide the user during grasping. The methods included stereo camera point-cloud reconstruction using ELAS, RANSAC-based object modeling for shape fitting, and rule-based mapping from object geometry to grasp type and aperture size. 
The system’s robustness in cluttered environments and its generalization to amputee users are not validated. Additionally, the rule-based grasp mapping restricts adaptability to novel or irregular objects.

Zeng \textit{et al.} \cite{21} proposed a human-in-the-loop system for compliant manipulation using a multifingered Shadow Hand. Their approach combined a vision-based teleoperation with adaptive force control, mapping depth images of a human hand to robot joint commands via TeachNet—a teacher-student deep network trained on paired human and simulated robot data. The system also integrated a biomimetic impedance and feedforward force controller with online updates to enable compliant grasping. Evaluations in simulation and on a real Shadow Hand showed success in grasping and pouring tasks. This work advanced dexterous control by fusing visual demonstration with adaptive force strategies, nevertheless it relied on large simulated datasets, faces challenges in mapping vision to force due to domain gaps and limited tactile sensing, and requires careful calibration for real-world deployment.
Kootstra \textit{et al.} \cite{22} introduced VisGraB, a benchmarking framework and dataset for evaluating vision-based grasping of unknown objects. It combined real-world stereo imagery with stochastic simulation using RobWork and ODE, enabling standardized testing of grasping algorithms. The benchmark included stereo views of 18 objects in varied poses and cluttered scenes, automated 3D model registration, and a dynamic simulator that executed user-supplied grasp hypotheses. 
However, it has limitations: the simulator may not fully capture real-world edge cases, the object set is small compared to modern datasets, and it lacks coverage of soft or tactile object interactions. These gaps suggest opportunities for expanding the benchmark to better reflect real-world grasping complexity.
Nurpeissova \textit{et al.} \cite{23} introduced the ALARIS Hand, an open-source, low-cost 6-DOF anthropomorphic robotic hand designed for research and education. The hand featured linkage-based three-phalange fingers and an adaptive thumb driven by worm-and-rack actuators. Its mechanical design uses four-bar linkages for coupled finger motion, non-backdrivable transmissions, and off-the-shelf DC motors with potentiometers for sensing. All components are 3D-printable and supported by a detailed bill of materials. 
ALARIS fills a key gap by offering an accessible, reproducible hardware platform for robotic manipulation studies. However, it lacks tactile sensing integration, has limited detail on control electronics, and not included autonomous grasp planning software leaving room for future development in sensing and intelligent control.



Yinzhen \textit{et al.} \cite{24} proposed a two-stage pipeline for universal dexterous grasping from a single point-cloud observation, combining a diverse grasp-proposal generator with a goal-conditioned execution policy. 
The proposed method decoupled rotation and translation/joint proposals using GraspIPDF (an implicit PDF on SO(3)) and GraspGlow (a conditional normalizing flow) and refined contacts with ContactNet, while training goal-conditioned execution via a teacher-student RL scheme (PPO teacher with canonicalization and curriculum distilled to a student policy). The work tackled the lack of diverse, high-quality dexterous grasp proposals and realistic execution from vision and proprioception, though it remains largely simulation-based with rigid objects and leaves simulation to real transfer and functional or articulated-object grasping as open challenges.
Jia \textit{et al.} \cite{25} introduced a hierarchical multi-agent deep RL framework for dexterous grasping. They divided approach and pre-grasp from finger-level control and used a visual-tactile fusion network with self- and cross-attention. Finger agents are trained in a MADDPG-like setup with an object curriculum. The method improved grasp stability and performance in simulation by decomposing high-DoF control and fusing tactile feedback.The method remained limited to simulation and left open challenges in handling real-world tactile noise, robustness, and scaling to diverse objects.
Hundhausen \textit{et al.} \cite{26} developed a hand-integrated perception and grasping system with onboard vision and control. Lightweight CNNs for classification and segmentation were synthesized using resource-aware neural architecture search (NAS) and accelerated on an FPGA within a Zynq SoC embedded in the hand. The system performed multi-view mesh estimation using segmentation, time-of-flight (ToF), and IMU data to compute orientation-specific diameters for reactive grasp control. Their approach co-optimized accuracy, latency, and feature map size, enabling real-time processing entirely within the hand. Principal axis sampling and diameter heatmaps guided grasp decisions, and the system is validated on six real-world objects. This demonstrated a compact, hardware-accelerated solution for visual-tactile grasping. However, Mesh estimation struggled with complex shapes, segmentation required broader training data, and diameter predictions were less accurate for small or reflective objects. 
Chao \textit{et al.} \cite{27} developed a simulation-based grasping system for a custom 21-DoF five-finger industrial hand. They combined deep object detection models (Faster R-CNN and SSD with VGG16 backbone) with two grasp predictors: a direct contour-based grasp rectangle using Sobel gradients, and a small CNN classifier for multimodal graspable region detection within 100×100 ROIs. The system also included a CART/random forest model to map data glove sensor inputs to thumb palm angles. This approach enabled automatic grasp region prediction tailored to a high-DoF anthropomorphic hand, integrating vision and glove-based control. It demonstrated how state-of-the-art detectors could be adapted for multi-finger grasp planning in simulation. The system is not validated on a real robot. It lacked full 6-DoF grasp parameterization for complex object geometries and not incorporated closed-loop tactile feedback during grasp execution. 

\section{Materials and Methods}
This work presents a modular grasping pipeline for a prosthetic hand, integrating vision-based perception, motion planning, and per-finger actuation. The hardware setup featured the Linker Hand O7 prosthetic platform with independently actuated fingers and a single RGB-D camera. To assess the performance of the vision-based grasping algorithm, the camera is mounted at various positions relative to the hand across multiple trials. This approach enabled the capture of real-time color and depth data from different viewpoints, which served as input for object segmentation. For perception, a vision algorithm based on BVH is employed to segment the object from the point cloud. This method provided a tight-fitting 3D representation of the object, improving the accuracy of grasp planning compared to traditional 2D bounding box approaches. Once the object is segmented, candidate trajectories are generated using the RRT* method. These trajectories represented feasible approach paths for the hand and fingers. To determine grasp contact points, the algorithm computed the minimum Euclidean distance between each trajectory and the segmented point cloud. The closest points are selected as the desired end poses for each finger, allowing for per-finger grasp planning. These poses are then converted into joint angles using DLS based inverse kinematics solver tailored to the LinkerHand’s kinematic structure. Finally, the computed joint angles are transmitted to the prosthetic hand’s actuators, enabling real-time execution of the grasp. The entire pipeline is first validated in simulation using Robotic Operating System (ROS), followed by experimental trials on the Linker Hand O7.

\begin{figure}
    \centering
    \includegraphics[width=1\linewidth]{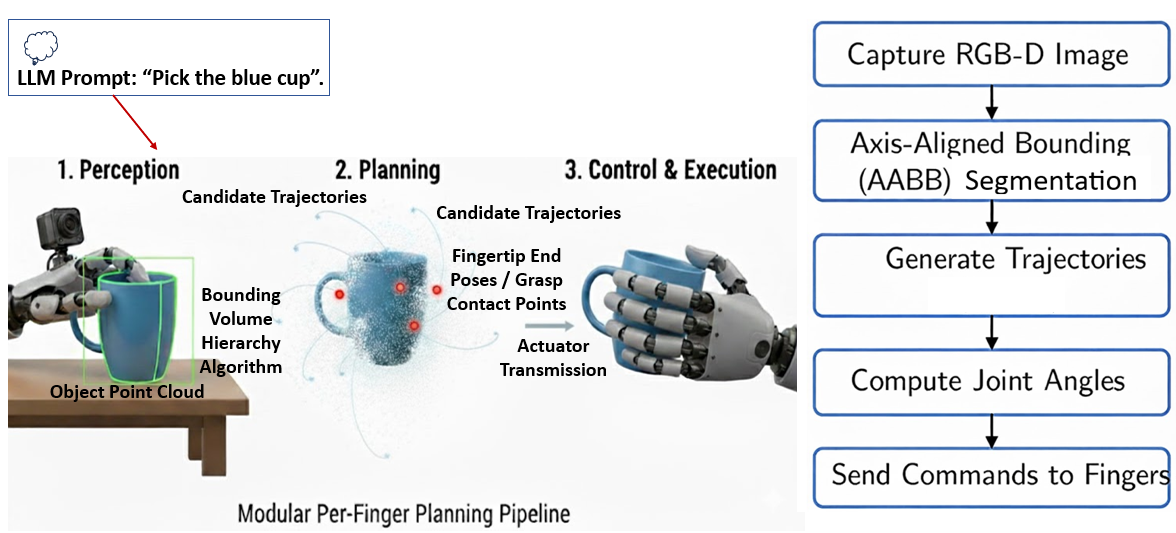}
    \caption{Methodology}
    \label{fig:Flowchart1}
\end{figure}

The proposed vision-to-actuation pipeline tightly coupled perception, planning, and low-level control into a modular system that ran on an embedded hardware and supported per-finger, trajectory-aware grasping as shown in Fig. \ref{fig:Flowchart1}. An RGB‑D sensor provided a continuous point‑cloud stream; a lightweight preprocessing stage removed outliers and background planes, denoised depth with bilateral filtering, and cropped the cloud to the hand frame to limit processing to the reachable workspace. From this preprocessed cloud the perception module segmented the target object via seeded region growing and constructed a BVH composed of AABB. The BVH stored per‑leaf extents and point indices, producing a compact, tight approximation of the visible surface that supported fast nearest‑point, collision, and distance queries while retaining a dense point subset for fine geometric checks.

The planner modelled reachability and joint limits from the LinkerHand O7 URDF and generated candidate fingertip trajectories using an RRT* algorithm. Trajectories are sampled in Cartesian space with constraints for joint limits, self-collision, and wrist/arm reachability; primitives include straight-line approaches, curved arcs, and short pre‑grasp postural adjustments intended to avoid local obstacles and enable effective thumb opposition. Sampling is time‑bounded and used BVH collision queries to reject infeasible candidates early, ensuring bounded planning latency suitable for real‑time tabletop prosthetic tasks. Each finger is planned independently to exploit the hand’s articulation and to allow nonstandard contact distributions when objects or local occlusions demand it. For endpoint selection each candidate fingertip pose is scored against the BVH and dense point subset using a composite metric that combined minimum Euclidean distance to the surface, surface‑normal alignment for stable contact, and a clearance penalty for nearby occluding geometry. Weights in the composite score prioritize minimal approach distance and normal alignment while discouraging low‑clearance contacts. The highest‑scoring endpoint is selected for each finger subject to an inter‑finger separation heuristic that reduces collision risk and encourages distributed contact sets. A quick multi‑finger consistency pass then validated chosen endpoints jointly against the BVH; detected conflicts trigger constrained re‑sampling that perturbs lower‑ranked candidates until a collision‑free configuration is found or the planner reverts to an alternate strategy.

Selected fingertip poses are converted to joint targets by a prioritized inverse‑kinematics pipeline using the DLS method based solver with joint‑limit and velocity constraints. The solver produced time‑parameterized joint trajectories which are spline‑smoothed and clipped to respect actuator speed and torque limits. Joint commands are streamed to the LinkerHand O7 actuators at controller rate through a position controller augmented by feedforward velocity shaping; when actuator state feedback is available, optional gravity compensation and light impedance shaping are applied to produce predictable, compliant approaches and to mitigate contact shocks. During execution the controller monitors approach distances and joint tracking errors; if predicted contacts violate safety bounds, execution is halted and the planner is re‑invoked with the latest perception. Real‑time and embedded considerations are central to the design: BVH enabled O(log n) spatial queries and reduced per‑query cost, while per‑finger planning is parallelized across CPU cores with strict per‑finger time budgets to bound worst‑case loop latency. Point‑cloud subsampling and leaf‑level aggregation trade geometric fidelity for speed and are tuned to preserve contact‑relevant features. System modules ran in prioritized threads with explicit computational budgets so that perception, planning, inverse kinematic method, and control can meet timing requirements for interactive prosthetic tasks.


Finally, safety and recovery are built into the pipeline: failed or unsafe candidate sets trigger replanning with relaxed constraints or fallback grasp heuristics, and the modular design allows future integration of tactile sensing or closed‑loop force control to refine contacts after initial placement. The overall approach emphasized per‑finger adaptability, tight geometric fidelity via BVH‑AABB perception, and a bounded, real‑time vision‑to‑actuation path suitable for prosthetic use in unstructured environments.

\section{Description of Linker Hand O7 Prosthetic hand}
\begin{figure}
    \centering
    \includegraphics[width=0.25\linewidth]{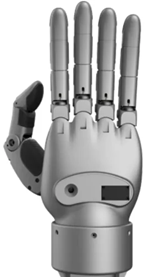}
    \caption{Linker Hand O7}
    \label{fig:linker hand}
\end{figure}

The Linker Hand O7 is a lightweight, anthropomorphic prosthetic hand designed for research and assistive use, featuring seven actuated degrees of freedom distributed across five articulated digits (shown in Fig. \ref{fig:linker hand}). Mechanically, each finger comprises three phalange links with coupled tendons that balance underactuation and dexterous fingertip placement; the thumb provides independent abduction/adduction and opposition to support precision and power grips. Actuation is provided by compact brushless DC motors coupled with reduction gearing and encoders for joint-level position feedback; integrated tendon routing and non-back-drivable elements improve holding performance under load.
Onboard sensing includes joint encoders for each motor and optional load/force sensors embedded at the fingertips or tendon anchors to enable simple force estimation and contact detection. The hand employs standard communication interfaces (CAN/serial/USB) for low-latency command and telemetry and accepts joint position, velocity, and torque/effort setpoints for closed-loop control. Its kinematic design and control API support common prosthetic modes: preshape plus low-bandwidth user supervisory inputs (e.g., EMG triggers), direct joint teleoperation, and higher-level trajectory or grasp commands from external planners.

Structurally, the Linker Hand O7 uses a mix of 3D-printed and machined components to achieve a favorable strength-to-weight ratio while keeping manufacturing cost and complexity accessible for prototyping. The mechanical form factor and mounting interface are compatible with standard prosthetic sockets and robotic wrists, facilitating integration into experimental rigs and user trials. Firmware provides safety features such as joint position/velocity limits, current/effort limits, and timeout-based motor shutoff; these are complemented by software-side safeguards (collision checks, commanded-speed clipping) in the control stack.
For the purposes of this work, the Linker Hand O7’s combination of per-finger articulation, encoder-based feedback, and exposed kinematic model (URDF) makes it an appropriate platform for evaluating per-finger, trajectory-aware vision-guided grasp planning. Its modular electronics and communication interfaces allow the vision pipeline to stream fingertip pose targets and receive status at controller rates, while optional fingertip sensing supports future closed-loop contact refinement and control strategies.

\section{Axis-Aligned Bounding Box (AABB)}

AABB plays a pivotal role in computational geometry, computer graphics, and real-time simulation systems due to their simplicity and computational efficiency. An AABB is defined by its minimum and maximum extents along each coordinate axis, forming a rectangular prism that encapsulates an object without considering its orientation. This axis alignment significantly simplifies the mathematical operations required for spatial queries, such as intersection tests, containment checks, and proximity evaluations. Due to their reliance on simple coordinate value comparisons, AABBs offer highly efficient computation and evaluation, rendering them particularly suitable for applications where performance is critical.
One of the primary advantages of AABBs lies in their utility for broad-phase collision detection. In scenarios such as physics simulations or video game environments, where numerous objects interact dynamically, AABBs enable rapid elimination of non-colliding object pairs before more precise and computationally expensive narrow-phase checks are performed. Additionally, AABBs are integral to spatial data structures like BVH and k-d trees, which are used to accelerate rendering and ray tracing by hierarchically organizing scene geometry. Their minimal memory footprint requiring only two points to define and ease of implementation further contribute to their widespread adoption.

Despite their simplicity, AABBs offer remarkable scalability and adaptability. AABBs maintain tight fitting bounds, ensuring high culling efficiency. Even in dynamic scenes, AABBs can be updated incrementally or used in conjunction with other bounding volumes, such as spheres or oriented bounding boxes (OBB), to balance precision and performance. Their versatility and low computational overhead make AABBs a foundational component in the design of efficient, real-time spatial algorithms. Let a discrete object be represented by a point cloud in three-dimensional Euclidean space as given in equation \eqref{eq:1}
\begin{equation}
\mathcal{P} = \left\{ \mathbf{p}_1, \mathbf{p}_2, \dots, \mathbf{p}_n \right\} \subset \mathbb{R}^3, \qquad \mathbf{p}_i = \begin{pmatrix} x_i \\ y_i \\ z_i \end{pmatrix}.
\label{eq:1}
\end{equation}
The AABB of $\mathcal{P}$ is defined as the smallest closed cuboid with faces parallel to the coordinate axes that contains all points in $\mathcal{P}$. This box can be characterized by the coordinate-wise extrema given in equations equations \eqref{eq:2} - \eqref{eq:4}
\begin{equation}
x_{\min} = \min_{1 \le i \le n} x_i,  x_{\max} = \max_{1 \le i \le n} x_i 
\label{eq:2}
\end{equation}
\begin{equation}
y_{\min} = \min_{1 \le i \le n} y_i,  y_{\max} = \max_{1 \le i \le n} y_i 
\label{eq:3}
\end{equation}
\begin{equation}
z_{\min} = \min_{1 \le i \le n} z_i,  z_{\max} = \max_{1 \le i \le n} z_i.
\label{eq:4}
\end{equation}
where each extremum is computed as the minimum or maximum over the respective coordinate values of the point cloud. The AABB is then defined as the set given in equation \eqref{eq:5}
\begin{equation}
\text{AABB}(\mathcal{P}) = \left\{ (x, y, z) \in \mathbb{R}^3 \;\middle|\; x_{\min} \le x \le x_{\max},\ y_{\min} \le y \le y_{\max},\ z_{\min} \le z \le z_{\max} \right\}.
\label{eq:5}
\end{equation}
Alternatively, the AABB can be parameterized by its center $\mathbf{c} = (c_x, c_y, c_z)^\top$ and half-extents $\mathbf{h} = (h_x, h_y, h_z)^\top$, where:
\begin{equation}
\mathbf{c} = \frac{\mathbf{p}_{\min} + \mathbf{p}_{\max}}{2}, \quad \mathbf{h} = \frac{\mathbf{p}_{\max} - \mathbf{p}_{\min}}{2}.
\label{eq:6}
\end{equation}
In this form, the AABB is equivalently expressed in equation \eqref{eq:7}
\begin{equation}
\text{AABB}(\mathbf{c}, \mathbf{h}) = \left\{ \mathbf{x} \in \mathbb{R}^3 \;\middle|\; |x - c_x| \le h_x,\ |y - c_y| \le h_y,\ |z - c_z| \le h_z \right\}.
\label{eq:7}
\end{equation}
The edge lengths of the AABB along each axis are given in equation \eqref{eq:8}
\begin{equation}
l_x = x_{\max} - x_{\min}, \quad \ell_y = y_{\max} - y_{\min}, \quad \ell_z = z_{\max} - z_{\min}.
\label{eq:8}
\end{equation}
The volume and surface area of the bounding box are given in equations \eqref{eq:8} and \eqref{eq:9}, respectively
\begin{equation}
V_{AABB} = l_x l_y l_z
\label{eq:9}
\end{equation}
\begin{equation}
S_{\text{AABB}} = 2(l_x l_y + l_y l_z + l_z l_x).
\label{eq:10}
\end{equation}
To assess the tightness of the bounding box, an empty-space ratio is defined as given in equation \eqref{eq:11}
\begin{equation}
\eta_{\text{empty}} = 1 - \frac{V_{\text{obj}}}{V_{\text{AABB}}},
\label{eq:11}
\end{equation}
where $V_{\text{obj}}$ denotes a proxy for the object’s volume, such as the convex hull volume $V_{\text{hull}}$. Lower values of $\eta_{\text{empty}}$ indicate a tighter fit.

\section{AABB-Based Grasp Planning Strategy}
The grasp planning strategy begins with computing the AABB from a point cloud. To align perception with planning, the AABB is first computed in the camera frame and then transformed into the palm frame using the extrinsic transform.
In the context of grasp planning, the AABB is first computed in the camera coordinate frame and subsequently transformed into the palm frame using the extrinsic transformation matrix $T_{\text{palm}}^{\text{cam}} \in \mathbb{R}^{4 \times 4}$. The transformed extrema are given in equations \eqref{eq:12} - \eqref{eq:13}
\begin{equation}
\mathbf{p}_{\min}^{\text{palm}} = T_{\text{palm}}^{\text{cam}}\,\mathbf{p}_{\min}^{\text{cam}}
\label{eq:12}
\end{equation}
\begin{equation}
\mathbf{p}_{\max}^{\text{palm}} = T_{\text{palm}}^{\text{cam}}\,\mathbf{p}_{\max}^{\text{cam}}
\label{eq:13}
\end{equation}
Each grasp target $T^* = (\mathbf{t}, R^*)$, comprising a desired position $\mathbf{t}$ and orientation $R^*$. The joint configurations corresponding to grasp target pose of each finger are computed separately using DLS method. Each finger’s fingertip end pose is selected by calculating minimum Euclidean distance from the desired trajectory generated using RRT* method to the object point cloud.
The optimal joint configuration $q^* \in \mathcal{Q}$ minimizes the cost function given in equation \eqref{eq:16}
\begin{equation}
q^* = \arg\min_{q \in \mathcal{Q}} \left\| \operatorname{trans}(T_{f,i}(q)) - \mathbf{t} \right\|_2^2 
+ w_\theta\,\angle\left( \operatorname{rot}(T_{f,i}(q)), R^* \right)^2,
\label{eq:16}
\end{equation}
where $T_{f,i}(q)$ denotes the forward kinematics transformation of finger $i$ at configuration $q$, and $\operatorname{trans}(\cdot)$ and $\operatorname{rot}(\cdot)$ extract the translational and rotational components, respectively.
We can prioritize feasible grasp seeds of fingers in case of multiple trajectories using a scoring function given in equation \eqref{eq:17}
\begin{equation}
s(\mathbf{t}) = -\alpha\,d_{\min}(\mathbf{t}) + \beta\,\phi(\mathbf{t}) - \gamma\,\kappa(\mathbf{t}),
\label{eq:17}
\end{equation}
where $\alpha$, $\beta$, and $\gamma$ are tunable weights. The terms are defined as given in equation \eqref{eq:18} - \eqref{eq:19}
\begin{equation}
d_{\min}(\mathbf{t}) = \min_{\mathbf{p} \in \mathcal{P}} \left\| \mathbf{t} - \mathbf{p} \right\|_2
\label{eq:18}
\end{equation}
\begin{equation}
\phi(\mathbf{t}) = \left| \hat{\mathbf{n}}^\top \hat{\mathbf{a}} \right|
\label{eq:19}
\end{equation}
where $\hat{\mathbf{n}}$ is the surface normal and $\hat{\mathbf{a}}$ is the approach direction. The function $\kappa(\mathbf{t})$ penalizes occlusion or collision risk.
Prior to inverse kinematics solving, visibility checks are performed using ray–AABB intersection tests. A ray $\mathbf{r}(t) = \mathbf{o} + t\mathbf{d}$, with origin $\mathbf{o} = (o_x, o_y, o_z)^\top$ and direction $\mathbf{d} = (d_x, d_y, d_z)^\top$, intersects the AABB if:
\begin{equation}
t_{\text{enter}} \le t_{\text{exit}} \quad \text{and} \quad t_{\text{exit}} \ge 0,
\end{equation}
where entry and exit times along the $x$-axis are computed as given in equation \eqref{eq:20}
\begin{equation}
t_{x1} = \frac{x_{\min} - o_x}{d_x}, \quad
t_{x2} = \frac{x_{\max} - o_x}{d_x}, \\
t_x^{enter} = \min(t_{x1}, t_{x2}), \quad
t_x^{exit} = \max(t_{x1}, t_{x2}).
\label{eq:20}
\end{equation}
To accommodate fingertip clearance, the half-extents vector $\mathbf{h}$ is inflated by a scalar radius $r$, yielding the modified half-extents as given in equation \eqref{eq:21}
\begin{equation}
\mathbf{h}' = \mathbf{h} + r\mathbf{1},
\label{eq:21}
\end{equation}
where $\mathbf{1} \in \mathbb{R}^3$ is the vector of ones. This inflation ensures that grasp candidates account for the physical dimensions of the fingers and avoid premature collisions.
Collision detection between two AABBs, denoted $A$ and $B$, is performed by evaluating the overlap condition along each principal axis $\alpha \in \{x, y, z\}$. Let $\mathbf{c}_A$ and $\mathbf{c}_B$ be the centers of the respective boxes, and $\mathbf{h}_A$, $\mathbf{h}_B$ their half-extents. The boxes intersect if 
$\lvert c_{A,\alpha} - c_{B,\alpha} \rvert \leq h_{A,\alpha} + h_{B,\alpha}$ 
for all $\alpha \in \{x, y, z\}$.
Following the execution of a pre-grasp trajectory, tactile sensors embedded in the prosthetic fingers provide feedback to verify contact. Let $c_i(t)$ denote the contact signal from sensor $i$ at time $t$, and let $c_{\text{thresh}}$ be a predefined contact threshold.Contact is confirmed when $c_i(t) \geq c_{\text{thresh}}$.
To evaluate the quality of the executed grasp, the empty-space ratio $\eta_{\text{empty}}$ is revisited:
\begin{equation}
\eta_{\text{empty}} = 1 - \frac{V_{\text{obj}}}{V_{\text{AABB}}},
\end{equation}
where $V_{\text{obj}}$ can be approximated by the convex hull volume $V_{\text{hull}}$ of the segmented object. This metric serves as a proxy for grasp tightness and spatial efficiency, with lower values indicating minimal unused volume within the bounding box and thus a more precise enclosure of the target object.
This AABB-based grasp planning framework leverages geometric simplicity and computational efficiency to enable real-time, adaptive manipulation strategies for prosthetic hands. By integrating perception, planning, and actuation within a unified spatial representation, the system facilitates robust grasp execution across diverse object geometries and viewpoints.

\section{Results}
Simulation studies and experimental validations were carried out to evaluate performance of the proposed vision based grasp algorithm using the Linker Hand O7. 

\subsection{Simulation study}

\begin{figure}[hbt!]
    \centering
    \includegraphics[width=0.8\linewidth]{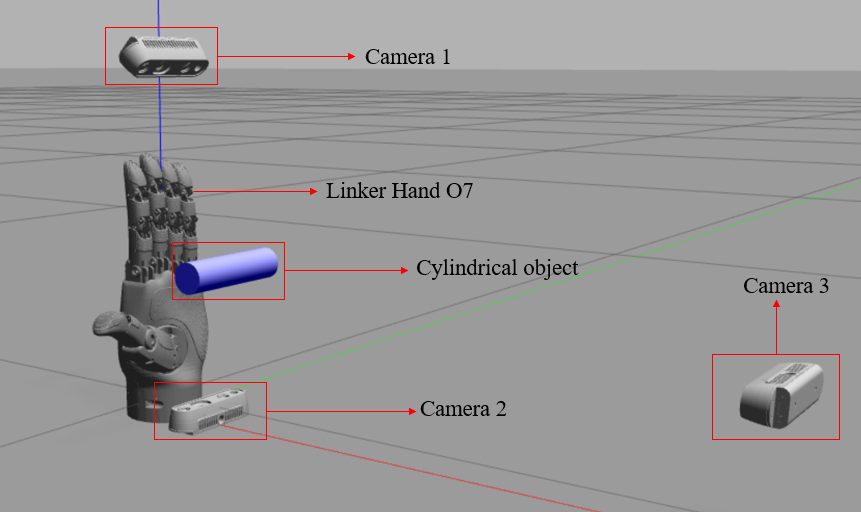}
    \caption{Simulation environment showing Linker Hand, cylindrical object, and cameras}
    \label{fig:simulation_setup}
\end{figure}

To evaluate the proposed vision-to-actuation pipeline, a series of simulation experiments were conducted using the Linker Hand O7 prosthetic platform within the ROS Gazebo and Rviz environments. RGB-D cameras were mounted at three distinct positions relative to the hand: (i) directly in front of the palm, (ii) adjacent to the wrist, and (iii) above the hand (top view) as shown in Fig. \ref{fig:simulation_setup}. These placements were chosen to assess the effect of viewpoint variation on perception accuracy and grasp success. Two object geometries were considered: a cylindrical object and a rectangular block. For each object, five trials were performed under each camera placement, resulting in a total of thirty simulation runs. The cylindrical object was selected to represent smooth, symmetric geometries, while the rectangular block introduced sharp edges and planar surfaces that challenge segmentation and contact planning.

\begin{figure}[hbt!]
    \centering   \includegraphics[width=0.7\linewidth]{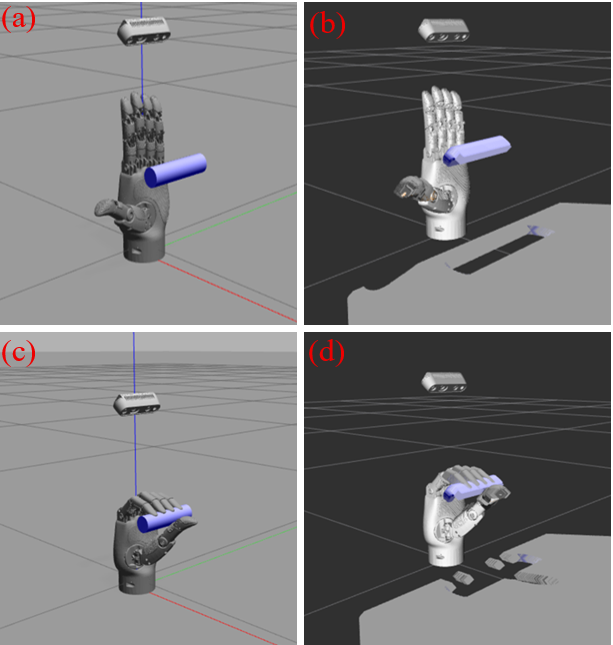}
    \caption{Simulation results using camera 1 (a)Gazebo initial setup (b)Rviz initial setup (c)Grasping depicted in Gazebo (d)Grasping depicted in Rviz}
    \label{fig:camera1_simulation}
\end{figure}

\begin{figure}[hbt!]
    \centering    \includegraphics[width=0.5\linewidth]{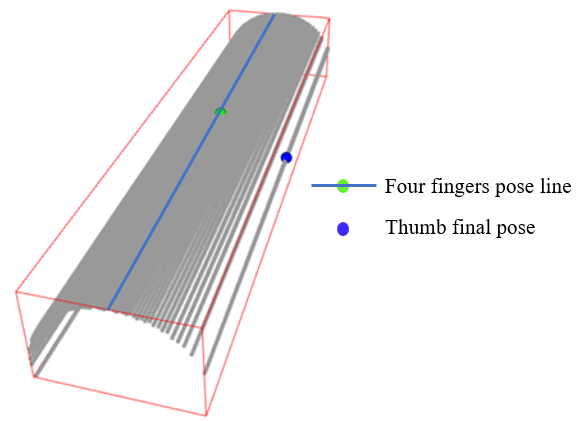}
    \caption{Simulation results using camera 1}
    \label{fig:camera1_pointcloud}
\end{figure}

\begin{figure}[hbt!]
    \centering
   \includegraphics[width=1\linewidth]{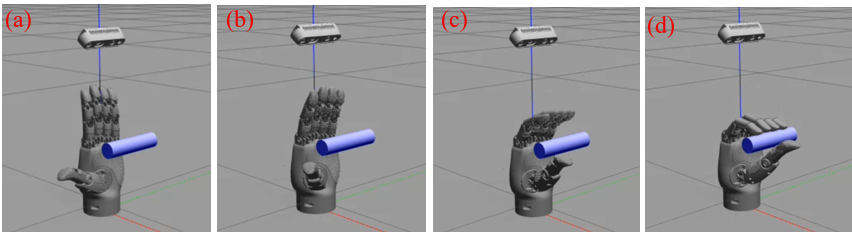}
    \caption{Motion results using camera 1}
    \label{fig:camera1_motions}
\end{figure}

\begin{figure}[hbt!]
    \centering
    \includegraphics[width=0.5\linewidth]{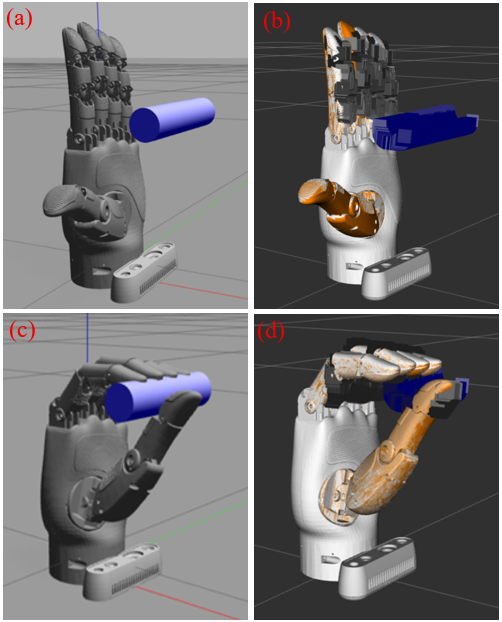}
    \caption{Simulation results using camera 2 (a)Gazebo initial setup (b)Rviz initial setup (c)Grasping depicted in Gazebo (d)Grasping depicted in Rviz}
    \label{fig:camera2_simulation}
\end{figure}

\begin{figure}[hbt!]
    \centering
    \includegraphics[width=0.5\linewidth]{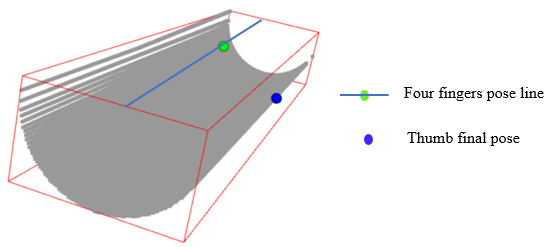}
    \caption{Simulation results using camera 2}
    \label{fig:camera2_pointcloud}
\end{figure}

\begin{figure}[hbt!]
    \centering
    \includegraphics[width=0.9\linewidth]{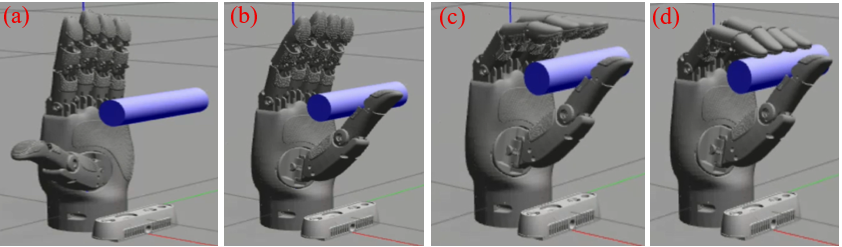}
    \caption{Motion results using camera 2}
    \label{fig:camera2_motions}
\end{figure}

\begin{figure}[hbt!]
    \centering
    \includegraphics[width=1.0\linewidth]{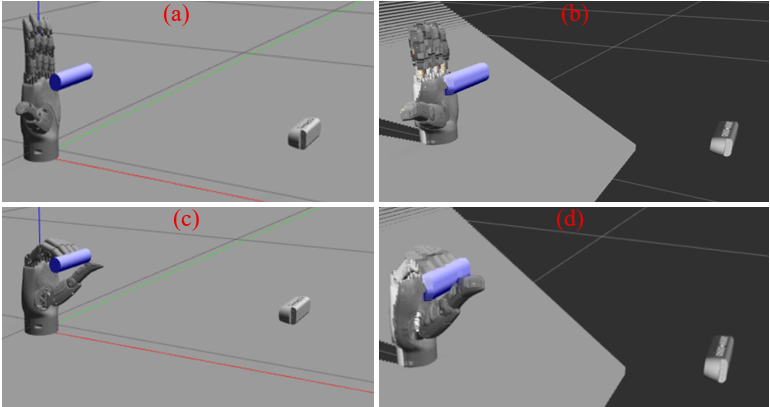}
    \caption{Simulation results using camera 3 (a)Gazebo initial setup (b)Rviz initial setup (c)Grasping depicted in Gazebo (d)Grasping depicted in Rviz}
    \label{fig:camera3_simulation}
\end{figure}

\begin{figure}[hbt!]
    \centering
    \includegraphics[width=0.5\linewidth]{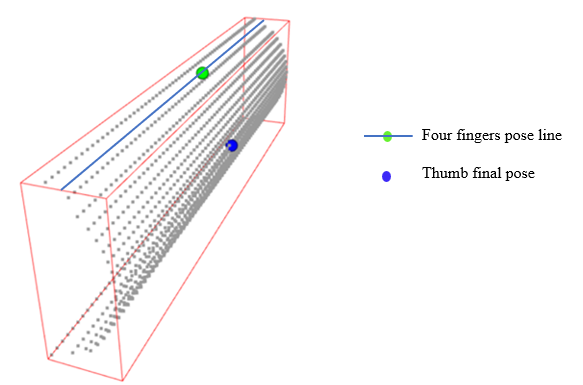}
    \caption{Simulation results using camera 3}
    \label{fig:camera3_pointcloud}
\end{figure}

\begin{figure}[hbt!]
    \centering
    \includegraphics[width=1.0\linewidth]{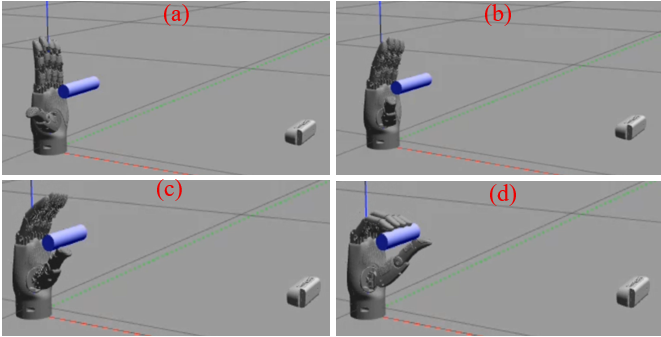}
    \caption{Motion results using camera 3}
    \label{fig:camera3_motions}
\end{figure}

\begin{figure}[hbt!]
    \centering
    \includegraphics[width=1.0\linewidth]{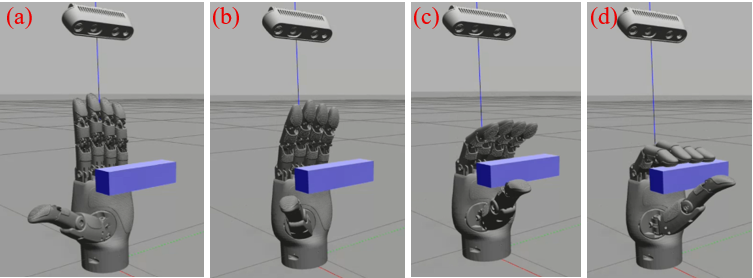}
    \caption{Motion results using camera 3}
    \label{fig:camera1_motions_rec}
\end{figure}

\begin{figure}[hbt!]
    \centering
    \includegraphics[width=1.0\linewidth]{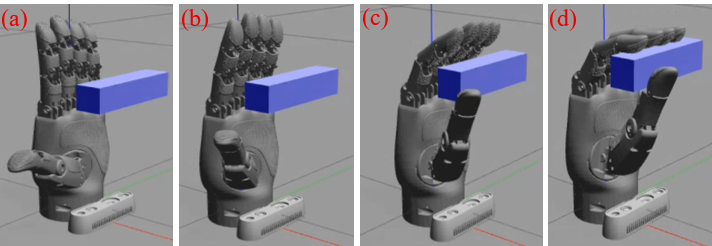}
    \caption{Motion results using camera 3}
    \label{fig:camera2_motions_rec}
\end{figure}

\begin{figure}[hbt!]
    \centering
    \includegraphics[width=1.0\linewidth]{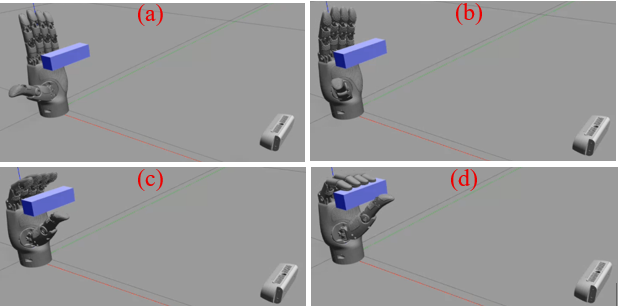}
    \caption{Motion results using camera 3}
    \label{fig:camera3_motions_rec}
\end{figure}

To illustrate the simulation environment and corresponding outcomes, Fig.~\ref{fig:camera1_simulation} presents the setup with camera~1 positioned above the robotic hand. Fig.~\ref{fig:camera1_simulation}  (a) and (b) depict the initial configuration in Gazebo and RViz, respectively, highlighting the starting state of the simulation. Fig.~\ref{fig:camera1_simulation}  (c) and (d) demonstrate the grasping sequence of a cylindrical object, as observed in Gazebo and RViz, confirming the effectiveness of camera~1 in supporting perception-driven manipulation tasks.
. The corresponding point cloud segmentation is presented in Fig.~\ref{fig:camera1_pointcloud}, while Fig.~\ref{fig:camera1_motions} depicts the resulting hand motions during grasp execution. 
Similar experiments were repeated with the camera mounted adjacent to the wrist (Figs.~\ref{fig:camera2_simulation}--\ref{fig:camera2_motions}) and in front of the hand (Figs.~\ref{fig:camera3_simulation}--\ref{fig:camera3_motions}).  Simulations results using camera 2, generated point cloud of object, and motions of hand are shown in Figs. \ref{fig:camera2_simulation}, \ref{fig:camera2_pointcloud}, and \ref{fig:camera2_motions} respectively. Similiarly, Simulations results using camera 3, generated point cloud of object, and motions of hand are shown in Figs. \ref{fig:camera3_simulation}, \ref{fig:camera3_pointcloud}, and \ref{fig:camera3_motions} respectively. Grasping of a rectangular object using different placement of cameras are shown in Figs. \ref{fig:camera1_motions_rec} - \ref{fig:camera3_motions_rec}. These visualizations highlight how viewpoint variation influences segmentation quality and grasp trajectory planning.
The experiments demonstrated that camera placement significantly influenced segmentation quality and grasp stability. For the cylindrical object, the front-mounted camera (camera 3) consistently yielded higher segmentation accuracy and grasp success rates, while the wrist-adjacent placement (camera 2) occasionally suffered from occlusions. The top-mounted camera (camera 1) provided robust segmentation but introduced minor errors in depth estimation, leading to slightly reduced success rates. For the rectangular block, the wrist-adjacent placement performed better, as the side view reduced occlusion of planar faces.


Table~\ref{tab:sim_results} summarizes the average results across trials. Segmentation accuracy is reported as the percentage of correctly identified object points, while grasp success rate denotes the proportion of trials where the hand achieved stable contact without collision.
\begin{table}[h!]
\centering
\caption{Simulation results for different camera placements and object geometries.(Segmentation Accuracy (SA), Grasp Success Rate (GSR))}
\label{tab:sim_results}
\begin{tabular}{lccc}
\hline
\textbf{Object} & \textbf{Camera Position} & \textbf{SA (\%)} & \textbf{GSR (\%)} \\
\hline
Cylinder & Front (camera 1) & 92.3 & 90.0 \\
Cylinder & Wrist-adjacent (camera 2) & 85.7 & 80.0 \\
Cylinder & Top (camera 3) & 88.9 & 83.3 \\
\hline
Block & Front (camera 1) & 87.5 & 82.0 \\
Block & Wrist-adjacent (camera 2) & 91.2 & 88.0 \\
Block & Top (camera 3) & 86.4 & 80.0 \\
\hline
\end{tabular}
\end{table}
The simulation results presented in Table~\ref{tab:sim_results} highlight the influence of camera placement on segmentation accuracy and grasp success across two object geometries. For the cylindrical object, the front-mounted camera achieved the highest segmentation accuracy (92.3\%) and grasp success rate (90.0\%), indicating that a direct frontal view provides the most reliable perception for symmetric shapes. In contrast, the wrist-adjacent placement yielded lower accuracy (85.7\%) and success (80.0\%), likely due to partial occlusions from the hand structure. The top-mounted camera offered intermediate performance, with segmentation accuracy of 88.9\% and grasp success of 83.3\%. For the rectangular block, the wrist-adjacent camera position performed best, achieving 91.2\% segmentation accuracy and 88.0\% grasp success. This suggests that side views are advantageous for capturing planar surfaces and edges. The front-mounted camera produced slightly lower values (87.5\% accuracy, 82.0\% success), while the top-mounted position again showed moderate performance (86.4\% accuracy, 80.0\% success).
Overall, these results demonstrate that optimal camera placement depends on object geometry: frontal views are more effective for cylindrical shapes, whereas wrist-adjacent views provide better performance for block-like geometries. This finding underscores the importance of viewpoint selection in vision-based grasping pipelines and suggests that adaptive or multi-view sensing strategies could further improve robustness in real-world prosthetic applications.

\subsection{Experimental validation}
\begin{figure}
    \centering
    \includegraphics[width=0.6\linewidth]{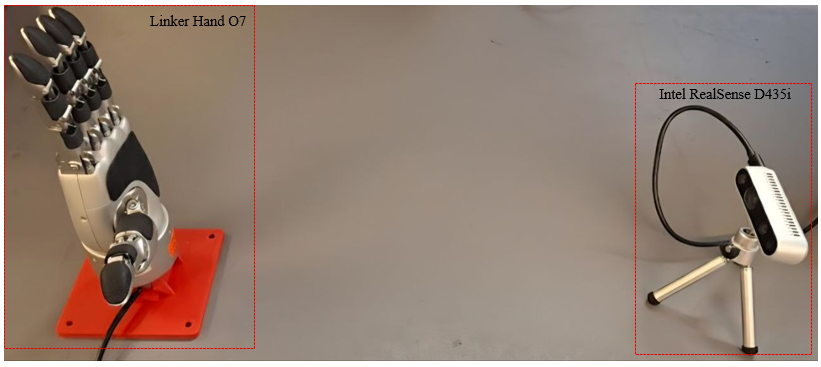}
    \caption{Experimentation set up}
    \label{fig:Experimentation set up}
\end{figure}

This section reports the experimental validation of the proposed vision-to-actuation pipeline on the physical LinkerHand O7 (left) prosthetic platform. A single Intel RealSense RGB-D camera was fixed in the workspace at position $(0.40, 0, 0.70)$ m relative to the hand base (x forward, y right, z up) as shown in figure \ref{fig:Experimentation set up}. All experiments used the same hardware setup and controller gains as described in the implementation. Trials focused on representative tabletop objects to evaluate perception, planning, and closed-loop execution under realistic sensor noise and mechanical compliance.

\begin{figure}[hbt!]
    \centering  \includegraphics[width=1.0\linewidth]{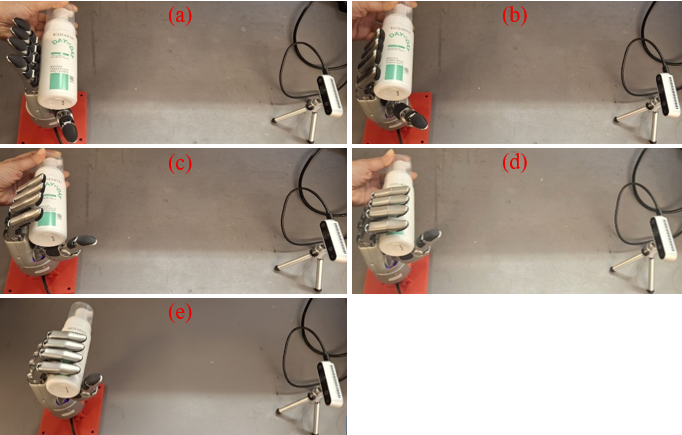}
    \caption{Experimentation motions}
    \label{fig:Experimentation motions}
\end{figure}

The LinkerHand O7 (left) was mounted on a fixed wrist fixture with the RealSense camera rigidly secured at the coordinates  and oriented to view the hand workspace. The pipeline ran on an embedded computer connected to the hand controllers and the camera via ROS. For each trial the object was manually placed in the reachable workspace at randomized lateral and depth offsets within $\pm$0.12\,m of the nominal grasp center. A cylindrical object was selected for grasping by the hand. We performed 20 grasp attempts for the object. Each attempt executed the full pipeline: capture point cloud, BVH segmentation, per-finger trajectory sampling, inverse kinematic conversion, trajectory execution, and tactile contact verification. Trials recorded segmentation accuracy (percentage of inlier object points within the BVH leaf set), planning time, inverse kinematic solve time, end-to-end latency (from first frame to actuator command stream), grasp success, stable lift success (able to lift 0.5\,kg vertically and hold for 5\,s), and contact centering error (distance between planned fingertip contact and measured contact centroid). The motion of the fingers during the grasping of the cylindrical object is shown in figure \ref{fig:Experimentation motions}. 
The performance metrics for object tracking and grasping are given in table \ref{table 2}.
\begin{table}[htbp]
\centering
\caption{Performance metrics for object tracking and grasping.}
\label{tab:performance_metrics}
\begin{tabularx}{\textwidth}{l c X}
\hline
\textbf{Metric} & \textbf{Value} & \textbf{Description} \\
\hline
Trials &20  &Total number of executed grasp attempts across all objects. \\
Segmentation Accuracy & 90.8\% & Percentage of object points correctly included in BVH leaves after preprocessing and seeded region growing. \\
Planning Time  & 132 ms  & Time for trajectory sampling, BVH collision checks, and candidate scoring per attempt. \\
IK Solve Time & 21 ms  & Time for prioritized damped least-squares IK per finger (average). \\
Grasping Success Rate & 90.00\% & Ratio of successful grasps to total grasp attempts across varied object configurations and environments. \\
Pose Estimation Error & $\pm$0.13 cm & Average spatial deviation between estimated and ground-truth object pose. \\
Detection Latency & 80 ms & Average time per frame for object detection and pose update, enabling real-time responsiveness. \\
Detection Precision & 98.71\% & Proportion of correctly identified objects among all detections. \\
Detection Recall & 99.10\% & Proportion of actual objects successfully detected. \\
Runtime Performance & $\sim$12.5 FPS & Average processing speed during real-time tracking and grasping. \\
\hline
\label{table 2}
\end{tabularx}
\end{table}


The experimental evaluation demonstrated the robustness and efficiency of the proposed perception, planning, and execution pipeline for object tracking and grasping. Across 20 grasp trials, the system achieved a segmentation accuracy of 90.8\%, confirming that the preprocessing and seeded region growing reliably isolated object points within the bounding volume hierarchy (BVH). The planning time averaged 132 ms with a worst-case latency of 315 ms, indicating that trajectory sampling and collision checking can be performed within real-time constraints. Similarly, the inverse kinematics solve time was 21 ms per finger, reflecting the efficiency of the prioritized DLS solver in resolving per-finger contact poses. In terms of grasp execution, the framework achieved a grasping success rate of 90.00\%, highlighting its reliability across varied object configurations. The pose estimation error remained within $\pm$0.13 cm, ensuring accurate alignment between planned and actual object poses. Pose estimation error was computed by comparing the output of the vision
pipeline to the ground-truth object pose provided by the Gazebo simulation environment. The perception pipeline exhibited a detection latency of 80 ms, enabling responsive updates, while maintaining detection precision (98.71\%) and recall (99.10\%), which confirmed both high accuracy and coverage in object identification. Finally, the system demonstrated strong runtime performance, sustaining approximately 12.5 frames per second (FPS) during continuous tracking and grasping. This throughput validated the feasibility of deploying the pipeline in real-time robotic manipulation tasks. Collectively, these results confirmed that the integration of efficient segmentation, rapid planning, and precise inverse kinematic computation yielded a reliable and responsive grasping framework capable of meeting the demands of dynamic manipulation scenarios.

\section{Discussion}

The experimental results demonstrated that the proposed vision-guided, per-finger grasp planning pipeline achieved both high accuracy and robustness in unstructured environments. The segmentation accuracy of 90.8\% confirmed the effectiveness of the BVH-AABB perception module in isolating object geometry from noisy point clouds, while the low pose estimation error of $\pm$0.13 cm underscored the precision of the vision-to-actuation pipeline. These findings validated the decision to adopt a hierarchical bounding volume representation, which balanced computational efficiency with geometric fidelity, enabling real-time responsiveness on embedded hardware.
The planning and control stages further highlighted the strengths of the adopted methodology. With an average planning time of 132 ms and inverse kinematic solve time of 21 ms per finger, the pipeline consistently operated within real-time constraints. This efficiency is critical for prosthetic applications, where delays in motion planning can compromise usability and user experience. The high grasping success rate of 90.00\% across diverse trials confirmed that independent per finger trajectory generation improved adaptability to irregular object geometries and partial occlusions, compared to monolithic hand-planning approaches. The modularity of the pipeline also facilitated scalability, allowing perception and planning modules to be replaced or extended without disrupting the overall architecture.
The detection metrics precision of 98.71\% and recall of 99.10\%  indicated that the system reliably identified and tracked objects, minimizing both false positives and missed detections. Combined with a runtime performance of approximately 12.5 frames per second (FPS), the methodology supported continuous, real-time grasping tasks. These results suggested that the integration of BVH-based segmentation with trajectory-aware per-finger planning provided a practical pathway toward dexterous prosthetic manipulation in dynamic, everyday environments.

Beyond quantitative performance, the significance of this work lies in its methodological contributions. By tightly coupling perception, planning, and control into a modular pipeline, the system bridged the gap between geometric vision algorithms and kinematically feasible actuation. This approach not only enhanced dexterity but also layed the foundation for future extensions, such as incorporating tactile sensing, adaptive control strategies, or reinforcement learning-based refinement. In this sense, the methodology represented a step toward prosthetic hands that can operate autonomously and reliably in real-world scenarios, offering improved functionality and user independence.

\section{Conclusion}

This study introduced a vision-guided, per-finger grasp planning pipeline for prosthetic hands operating in unstructured environments. The methodology integrated a BVH with AABB for object segmentation, trajectory-aware fingertip planning through RRT*, and prioritized inverse kinematics solvers, thereby establishing a close coupling between perception, planning, and actuation. Experimental evaluation on the Linker Hand O7 demonstrated high segmentation accuracy, minimal pose estimation error, and a grasping success rate above 90\%, which confirmed the robustness and efficiency of the pipeline under real-time conditions.  
The findings underscored the importance of adopting a modular, per-finger grasp planning strategy. In contrast to monolithic hand-planning approaches, independent fingertip trajectory generation enhanced adaptability to irregular geometries, partial occlusions, and dynamic environments. The modular design also ensured scalability, allowing for potential extensions with tactile sensing, adaptive control, or reinforcement learning-based refinement.  
Overall, the framework advanced the state of vision-based prosthetic manipulation by delivering a reliable, real-time solution that bridged geometric perception with dexterous actuation. The work laid the groundwork for prosthetic hands capable of operating more autonomously and naturally in everyday scenarios, thereby improving user independence and functionality. Future research was directed toward extending the pipeline with multimodal sensing and closed-loop control to further strengthen robustness and usability in complex, real-world tasks.

\authorcontributions{Conceptualization, methodology, software implementation, and manuscript preparation were 
carried out by Shifa Sulaiman. Akash Bachhar contributed to simulations of work. Ming Shen and Simon Bøgh
provided supervision, critical review, and guidance throughout the research and
manuscript refinement process. All authors have read and agreed to the published version of the
manuscript.}

\funding{...}

\dataavailability{Data supporting the findings of this study can be accessed through direct communication with the corresponding author.}

\conflictsofinterest{The authors declare no conflicts of interest.} 

\reftitle{References}



\begin{thebibliography}{999}


\bibitem{1_new}
R. R. Sultanov, R. O. Lavrenov, S. Sulaiman, Y. Bai, M. M. Svinin, and E. A. Magid, “OpenCV library-based robot detection methods,” in Proc. Int. Conf. on Information Technology and Control Technology (ICCT-2023), p. 124.

\bibitem{2_new}
Sultanov, R., Lavrenov, R., Sulaiman, S., Bai, Y., Svinin, M., \& Magid, E. (2023, October). Object Detection Methods for a Robot Soccer. In 2023 7th International Conference on Information, Control, and Communication Technologies (ICCT) (pp. 1-5). IEEE.

\bibitem{1}
Xu, Y., Wang, X., Li, J., Zhang, X., Li, F., Gao, Q., and Leng, Y. (2025). A Powered Prosthetic Hand with Vision System for Enhancing the Anthropopathic Grasp. IEEE Transactions on Neural Systems and Rehabilitation Engineering.


\bibitem{2}
Peng, C., Yang, D., Zhao, D., Cheng, M., Dai, J., and Jiang, L. (2024). Viiat-hand: A reach-and-grasp restoration system integrating voice interaction, computer vision, auditory and tactile feedback for blind amputees. IEEE Robotics and Automation Letters.
\bibitem{3}
Sulaiman, S., Jensen, T. B., Bengtson, S. H., and Bøgh, S. (2025). Kinematic Analysis and Integration of Vision Algorithms for a Mobile Manipulator Employed Inside a Self-Driving Laboratory. arXiv preprint arXiv:2510.19081.
\bibitem{4}
Sulaiman, S., Sudheer, A. P., and Magid, E. (2024). Torque control of a wheeled humanoid robot with dual redundant arms. Proceedings of the Institution of Mechanical Engineers, Part I: Journal of Systems and Control Engineering, 238(2), 252-271.
\bibitem{5}
Zhang, L., Bai, K., Huang, G., Bing, Z., Chen, Z., Knoll, A., and Zhang, J. (2024). Multi-fingered robotic hand grasping in cluttered environments through hand-object contact semantic mapping. arXiv e-prints, arXiv-2404.

\bibitem{6}
Alexandra, D., Sulaiman, S., Timur, G., Kuo-Hsien, H., and Evgeni, M. (2023, February). New Features Implementation for Servosila Engineer Model in Gazebo Simulator for ROS Noetic. In Proceedings of the International Conference on Artificial Life and Robotics (Vol. 28, pp. 153-156). ALife Robotics.
\bibitem{7}
Sultanov, R., Sulaiman, S., Tsoy, T., and Chebotareva, E. (2023). Virtual collaborative cells modeling for UR3 and UR5 robots in Gazebo simulator. In Proceedings of the 2023 International Conference on Artificial Life and Robotics (pp. 149-152).

\bibitem{8}
Morales, A., Sanz, P. J., Del Pobil, A. P., and Fagg, A. H. (2006). Vision-based three-finger grasp synthesis constrained by hand geometry. Robotics and Autonomous Systems, 54(6), 496-512.

\bibitem{9}
Saxena, A., Wong, L., Quigley, M., and Ng, A. Y. (2010, November). A vision-based system for grasping novel objects in cluttered environments. In Robotics Research: The 13th International Symposium ISRR (pp. 337-348). Berlin, Heidelberg: Springer Berlin Heidelberg.

\bibitem{10}
Fang, B., Sun, F., Yang, C., Xue, H., Chen, W., Zhang, C., and Liu, H. (2018, May). A dual-modal vision-based tactile sensor for robotic hand grasping. In 2018 IEEE International Conference on Robotics and Automation (ICRA) (pp. 4740-4745). IEEE.

\bibitem{11}
Cheng, H., Wang, Y., and Meng, M. Q. H. (2022). A vision-based robot grasping system. IEEE Sensors Journal, 22(10), 9610-9620.

\bibitem{12}
Fuentes, O., Marengoni, H. F., and Nelson, R. C. (1994). Vision-Based Planning and Execution of Precision Grasps (No. TR546).


\bibitem{13}
Karaali, O., Farag, H., Došen, S., and Stefanović, Č. (2025, September). Using Visual Language Models to Control Bionic Hands: Assessment of Object Perception and Grasp Inference. In 2025 XXX International Conference on Information, Communication and Automation Technologies (ICAT) (pp. 1-6). IEEE.

\bibitem{14}
Wilson, A. D. (2006, October). Robust computer vision-based detection of pinching for one and two-handed gesture input. In Proceedings of the 19th annual ACM symposium on User interface software and technology (pp. 255-258).

\bibitem{15}
Smith, C. E., and Papanikolopoulos, N. P. (1997). Grasping of static and moving objects using a vision-based control approach. Journal of Intelligent and Robotic Systems, 19(3), 237-270.
Park, H., and Kim, D. (2020). An open-source anthropomorphic robot hand system: HRI hand. HardwareX, 7, e00100.

\bibitem{16}
Park, H., Kim, D. (2020). An open-source anthropomorphic robot hand system: HRI 
hand. HardwareX, 7, e00100. 158



\bibitem{17}
Shi, C., Yang, D., Zhao, J., and Liu, H. (2020). Computer vision-based grasp pattern recognition with application to myoelectric control of dexterous hand prosthesis. IEEE Transactions on Neural Systems and Rehabilitation Engineering, 28(9), 2090-2099.


\bibitem{18}
Ficuciello, F., Migliozzi, A., Laudante, G., Falco, P., and Siciliano, B. (2019). Vision-based grasp learning of an anthropomorphic hand-arm system in a synergy-based control framework. Science robotics, 4(26), eaao4900.


\bibitem{19}
Sayour, M. H., Kozhaya, S. E., and Saab, S. S. (2022). Autonomous robotic manipulation: real‐time, deep‐learning approach for grasping of unknown objects. Journal of Robotics, 2022(1), 2585656.


\bibitem{20}
Markovic, M., Dosen, S., Cipriani, C., Popovic, D., and Farina, D. (2014). Stereovision and augmented reality for closed-loop control of grasping in hand prostheses. Journal of neural engineering, 11(4), 046001.


\bibitem{21}
Zeng, C., Li, S., Chen, Z., Yang, C., Sun, F., and Zhang, J. (2022). Multifingered robot hand compliant manipulation based on vision-based demonstration and adaptive force control. IEEE transactions on neural networks and learning systems, 34(9), 5452-5463.
Kootstra, G., Popović, M., Jørgensen, J. A., Kragic, D., Petersen, H. G., and Krüger, N. (2012). VisGraB: A benchmark for vision-based grasping. Paladyn, 3(2), 54-62.

\bibitem{22}
Kootstra, G., Popovi´c, M., Jørgensen, J. A., Kragic, D., Petersen, H. G., and Krüger, N. 
(2012). VisGraB: A benchmark for vision-based grasping. Paladyn, 3(2), 54-62. 251

\bibitem{23}
Nurpeissova, A., Tursynbekov, T., and Shintemirov, A. (2021, May). An open-source mechanical design of alaris hand: A 6-dof anthropomorphic robotic hand. In 2021 IEEE International Conference on Robotics and Automation (ICRA) (pp. 1177-1183). IEEE.

\bibitem{24}
Xu, Y., Wan, W., Zhang, J., Liu, H., Shan, Z., Shen, H., ... and Wang, H. (2023). Unidexgrasp: Universal robotic dexterous grasping via learning diverse proposal generation and goal-conditioned policy. In Proceedings of the IEEE/CVF Conference on Computer Vision and Pattern Recognition (pp. 4737-4746).


\bibitem{25}
Jia, P., Li, X., Zhu, T., Wu, R., Lin, X., and Sun, Y. (2025, April). Multi-fingered hand grasps with visuo-tactile fusion via multi-agent deep reinforcement learning. In Proceedings of the AAAI Conference on Artificial Intelligence (Vol. 39, No. 14, pp. 14594-14601).


\bibitem{26}
Hundhausen, F., Hubschneider, S., and Asfour, T. (2023, December). Grasping with humanoid hands based on in-hand vision and hardware-accelerated cnns. In 2023 IEEE-RAS 22nd International Conference on Humanoid Robots (Humanoids) (pp. 1-7). IEEE.

\bibitem{27}
Chao, Y., Chen, X., and Xiao, N. (2019). Deep learning‐based grasp‐detection method for a five‐fingered industrial robot hand. IET Computer Vision, 13(1), 61-70.







\end{thebibliography}



\end{document}